\documentclass{article}

\usepackage{microtype}
\usepackage{graphicx}
\usepackage{subfigure}
\usepackage{booktabs} % for professional tables

\usepackage{hyperref}

\usepackage[accepted]{icml2024}

\usepackage{amsmath}
\usepackage{amssymb}
\usepackage{mathtools}
\usepackage{amsthm}

\usepackage{enumitem}

\usepackage[capitalize,noabbrev]{cleveref}

\theoremstyle{plain}

\theoremstyle{definition}

\theoremstyle{remark}

\newcommand{\parag}[1]{\textbf{#1}}

\newcounter{displaybox}
\usepackage{tcolorbox}

\definecolor{myblue}{HTML}{4472c4}
\newcommand{\takeaway}[1]{\textcolor{myblue}{\textbf{Takeaway:} #1}}

\usepackage{breakurl}

\icmltitlerunning{Position: Why Tabular Foundation Models Should Be a Research Priority}

\begin{document}

\twocolumn[
\icmltitle{Position: Why Tabular Foundation Models Should Be a Research Priority}



\icmlsetsymbol{equal}{*}

\begin{icmlauthorlist}
\icmlauthor{Boris van Breugel}{cam}
\icmlauthor{Mihaela van der Schaar}{cam,turing}
\end{icmlauthorlist}

\icmlaffiliation{cam}{Department of Applied Mathematics and Theoretical Physics, University of Cambridge, UK}
\icmlaffiliation{turing}{Alan Turing Institute, London, UK}

\icmlcorrespondingauthor{Boris van Breugel}{bv292@cam.ac.uk}

\icmlkeywords{Machine Learning, ICML}

\vskip 0.3in
]



\printAffiliationsAndNotice{}  % leave blank if no need to mention equal contribution

\begin{abstract}
Recent text and image foundation models are incredibly impressive, and these models are attracting an ever-increasing portion of research resources. In this position piece we aim to shift the ML research community's priorities ever so slightly to a different modality: tabular data. Tabular data is the dominant modality in many fields, yet it is given hardly any research attention and significantly lags behind in terms of scale and power. \textbf{We believe the time is now to start developing tabular foundation models}, or what we coin a \textit{Large Tabular Model} (LTM). LTMs could revolutionise the way science and ML use tabular data: not as single datasets that are analyzed in a vacuum, but contextualized with respect to related datasets. The potential impact is far-reaching: from few-shot tabular models to automating data science; from out-of-distribution synthetic data to empowering multidisciplinary scientific discovery. We intend to excite reflections on the modalities we study, and convince some researchers to study large tabular models.
\end{abstract}

% overview
\section{Introduction} \label{sec:introduction}
% - incredible advances in ML for text and image
% - tabular data is lacking. 
% - tabular data has potential Humans are bad at tabular data

Let us start with the obvious: the recent progress in modelling text and image is incredibly impressive. It is not just the capabilities of these models that has grabbed the public imagination, it is also their seemingly ``creative'', human-like output; photorealistic images of horse-riding astronauts and poems in the style of Sylvia Plath. We acknowledge that text and image foundation models have large potentials for real-world good---from low-cost, tailored educational aids, to personalized medicine. Furthermore, these models are already showcasing how the use of ML need not be constrained to the ML community.
% go into past and emergence properties

\begin{figure}[hbt]
    \centering
    \includegraphics[width=\columnwidth]{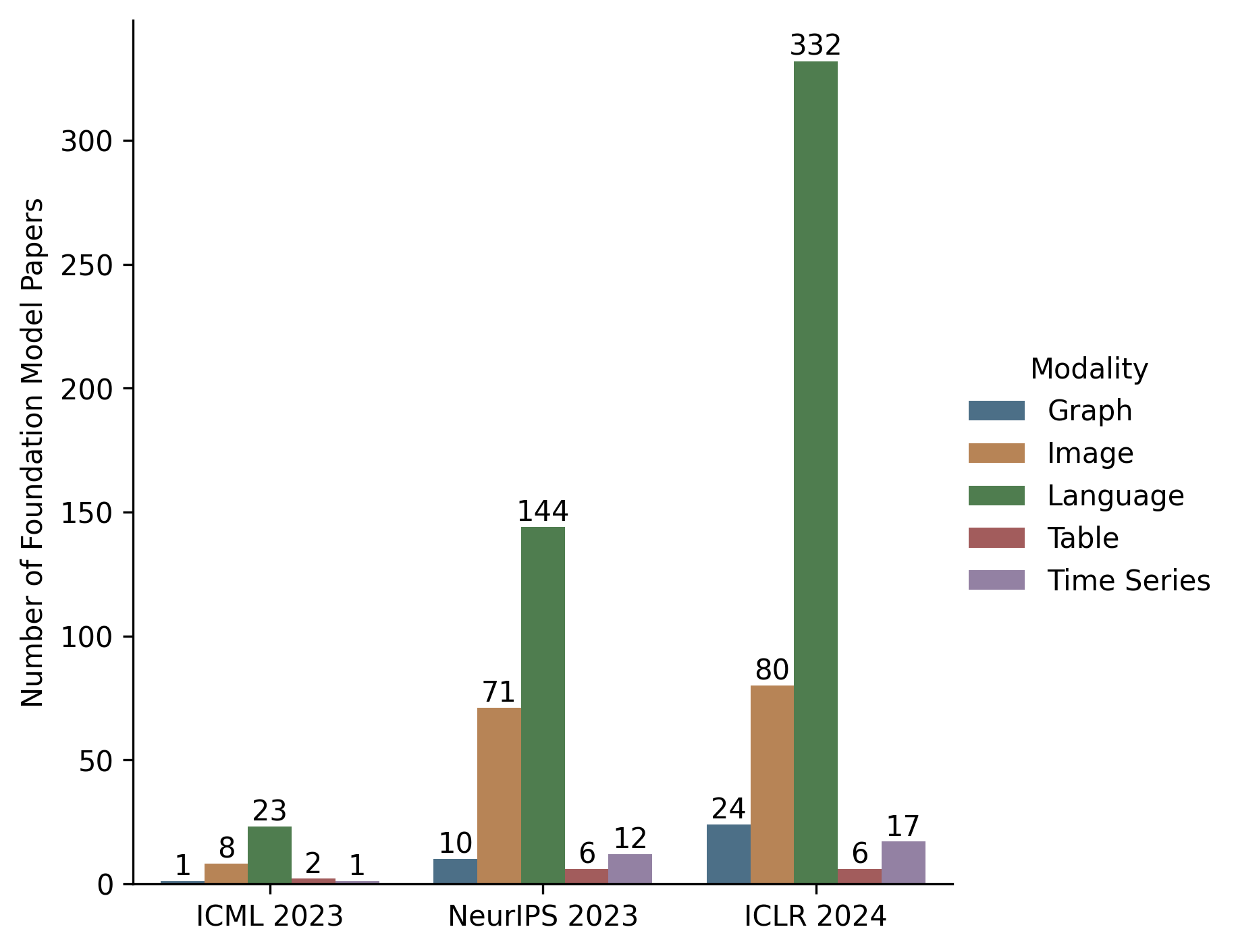}
    \vspace{-5mm}
    \caption{Representation of different modalities in foundation model research across recent ML conferences, roughly estimated as the number of accepted papers with abstracts containing keywords (see Appendix \ref{app:figure_details}). LLMs are booming and tabular data is heavily underrepresented.}
    \label{fig:conference_stats}
\end{figure}

The immense public interest in text and image modalities, as well as commercial incentives, may partly explain why the majority of foundation model (FM) research is focused on these modalities. The flip side is that other modalities receive hardly any attention from the ML research community (see Figure \ref{fig:conference_stats}). With this work, we prompt foundation model researchers to reflect on the modality they study, to consider alternative modalities, and to potentially redirect their attention. In particular, in this paper we will argue how \textit{tabular} foundation models have been almost entirely unexplored, yet in some domains present a potential for impact that is just as large, if not larger, than image and text models. We will refer to this class of models as Large Tabular Model (LTM).

% At the same time, the amount of freely available data is growing steadily. Foundation models in the image and text domains have already captured the popular imagination, showing what these vast quantities of data can achieve. In this work, we explore the potential and possible form of similar generative models, but for the tabular data domain. These large 
\parag{How to think about LTMs?}
LLMs are already widely used as tools, and vision FMs (e.g. \cite{saharia_photorealistic_2022}) can generate synthetic images or embed real images for downstream tasks. LTMs can play a similar role for (multimodal settings with) tabular data. For data scientists, LTMs could provide an invaluable tool for cleaning and preprocessing datasets; for finding relevant datasets (possibly from different domains or general knowledge-bases, e.g. Wikipedia tables); for augmenting existing datasets (e.g. few-shot generation of additional columns or performing SQL joins based on related data, e.g. adding a GDP column to a dataset with different countries); and for conducting automated (meta)-analyses. Just like vision FMs, generative LTMs could be used to generate synthetic data, and do so with less data than conventional generative models. Generating full, accurate synthetic datasets that can be used in-place of real data has many applications \cite{van_breugel_beyond_2023}: enable data sharing while maintaining privacy, reduce bias and improve representation of marginalized groups, simulate unseen domains, and augment the size of existing real data. Just like vision and text FMs, using the LTM's embedding of tabular rows (or full datasets) could be used for downstream tasks, including as input for prediction models or other foundation models.

\parag{Why tabular FMs have been overlooked.} We may wonder why LTMs have been overlooked so far. We hypothesize there are three main reasons: data, tabular ML difficulty, and human perception. First, until recently there has been a lack of large tabular metadatasets. These datasets may also be messy or not fit the typical ML problem setting (e.g. are not labelled for any specific task, or may require domain knowledge). Additionally, due to privacy or ownership concerns, some subfields (e.g. medical data) would not be represented in this data. Second, the difficulty of tabular ML can be discouraging. As discussed, tabular data baselines are often strong, and a new method may not consistently outperform it---possibly hindering publication as a result. Furthermore, there are some unique challenges (see desiderata \ref{sec:define}). Third, we believe humans' ``more natural" skills of interpreting text and vision may have played a large role. Many vision and text papers have as primary way of evaluation human judgement (e.g. realism of generated images or texts). This is a solution when metrics are unavailable or unreliable---e.g. the primary image generation metrics FID and IS have known problems \cite{liu_improved_2018,chong_effectively_2019,alaa_how_2022}---yet visual inspection is very hard for tabular data. At last, non-experts understand that StableDiffusion and ChatGPT generate impressively realistic outputs, and hence these models are featured heavily in and outside the ML research community. This in turn may have encouraged more people to study this topic compared to modalities like time-series and tabular.

\parag{Why care about the tabular domain.}
Let us highlight three main reasons why switching to tabular foundation models is worth your consideration. First, \textbf{tabular data is ubiquitous} in the real world \cite{borisov_deep_2022, shwartz-ziv_tabular_2022}---from electronic healthcare records \cite{fatima_survey_2017} to census data \cite{office_for_national_statistics_2021_2021}, from cybersecurity \cite{buczak_survey_2016} to credit scoring \cite{dastile_statistical_2020}, and from finance to natural sciences \cite{shwartz-ziv_tabular_2022}. Quantitative research relies on these datasets, which in turn progresses scientific knowledge and influences public policy. This practical importance is also reflected in its prominence in online data science competitions, e.g. Kaggle and KDD Cup focus primarily on tables \cite{kaggle_2017_2017, huang_tabtransformer_2020}. 

Second, despite the above \textbf{the tabular domain offers uniquely exciting, large, unsolved challenges for researchers}. In the past two decades, machine learning for tabular data has not progressed as resolutely compared to other modalities. Recent benchmarking papers \cite{gorishniy_revisiting_2021,borisov_deep_2022, shwartz-ziv_tabular_2022,grinsztajn_why_2022} all find that XGBoost \cite{chen_xgboost_2016}, a tree-based model, is still among the top performers for supervised learning on tabular data. The authors attribute many reasons to this, e.g. tabular data being discontinuous, containing heterogeneous and uninformative features \cite{grinsztajn_why_2022}, and data containing no spatial invariances that could inform a good prior (cf. convolutional nets for vision) \cite{borisov_deep_2022}. We will go into some of these further in Section \ref{sec:define}, but for now we would like to point to another, more obvious reason why tabular data research is lagging behind other modalities: the scale of data, models, and task complexity is vastly different than in the vision and text domain. In \cite{grinsztajn_why_2022}, the largest datasets are only 50,000 points, and for these datasets (Appendix A2) the difference between data-heavy transformer-based models and tree-based models becomes smaller. Similarly, \citet{borisov_deep_2022} find that for their largest dataset (which counts 11 million samples but just 27 features), the transformer-based model SAINT \cite{somepalli_saint_2021} outperforms tree-based models. Additionally, benchmark papers \cite{kadra_well-tuned_2021,gorishniy_revisiting_2021,mcelfresh_when_2023} do find settings in which neural nets seem to outperform XGBoost rather consistently, e.g. when the number of samples increases \cite{mcelfresh_when_2023}, the number of target classes increases \cite{gorishniy_revisiting_2021}, and when more modern regularization techniques are used for deep nets \cite{kadra_well-tuned_2021}. Furthermore, none of the discussed tabular benchmarking papers consider settings with multiple or truly large and diverse datasets---even though the power of modern vision and text models may well be attributed to their training on billions of images \cite{schuhmann_laion-5b_2022} or trillions of tokens \cite{touvron_llama_2023}. These observations indicate that the scaling advantages observed in other modalities \cite{bommasani_opportunities_2022}, may well hold in tabular data, yet these model sizes and data settings are simply not studied due to benchmarks consisting of primarily smaller datasets and models. This comes with another advantage: \textbf{developing SOTA LTMs is still within computational reach of many ML researchers}, cf. the economically-exclusionary cost of training modern LLMs.

Third, the impact of a foundation model for tabular data lies not just in the ubiquity of the modality, but also in \textbf{humans' intrinsic inability to parse tabular data effectively \emph{themselves}}. Humans are incredibly good at understanding text and image, and foundation models in the text and vision space have aimed to match this: to encapsulate fundamental understanding of abstract concepts, resembling human-like skills. In the tabular domain the power of FMs may be just as large---foundation models being able to reason about real-world distributions over different variables, and generalize to new relationships. For example, a foundation model trained on tables with features A and B, and B and C, might well be able to reason about the relationship between A and C. Relatively speaking, however, this power would be much higher compared to humans, as table parsing, data analyses, and computations are not an intrinsic skill of ours.\footnote{In fact, we expect many readers consider tables as abstract and ``dry'', and may well be the reason why it is so understudied. } We do not know what knowledge can be derived from combining wide, diverse tabular datasets, but the potential alone is exciting and worth exploring.

In a nutshell, designing the first generation of truly large, foundational tabular models could be \emph{an immensely rewarding and exciting opportunity} for many researchers.

\parag{Overview.} In this position piece, we start by contextualizing the term LTM with respect to other foundation models \cite{bommasani_opportunities_2022}, and discussing model and data requirements (Section \ref{sec:define}). In Section \ref{sec:related}, we will discuss the current research related to LTMs, and how it is still rather limiting. We continue exploring the applications of LTMs (Section \ref{sec:applications}), as well as adaptation challenges (Section \ref{sec:challenges}). At last, in Section \ref{sec:comparison} we return to our thesis and provide a head-to-head comparison of the impact of LTMs and LLMs along different dimensions, attempting to convince the ML research community to shift more attention to tabular.

\section{Large Tabular Model} \label{sec:define}
The term Foundation Model (FM) was first coined by \cite{bommasani_opportunities_2022}, denoting ``any model that is trained on broad data (generally using self-supervision at scale) that can be adapted (e.g., fine-tuned) to a wide range of downstream tasks; current examples include BERT \cite{devlin_bert_2019}, GPT-3 \cite{brown_language_2020}, and CLIP \cite{radford_learning_2021}.'' The three mentioned examples being LLMs reflect the initial dominance of LLMs in FM research and discourses. Though \cite{bommasani_opportunities_2022} write about FMs for other modalities (including image and multimodal), tabular FMs are missing from the 200+ page report. 

To draw a direct comparison with LLMs, we coin the term Large Tabular Model (LTM) for a tabular foundation model. In this paper, we focus on the simplest form of tabular data: single independent tables (cf. relational databases). Let us discuss how the foundation model definition moulds to tabular data, and what kind of model and data it necessitates.

\subsection{Model Requirements}
The two foundation model requirements, large-scale (self-supervised) training and adaptability, extent to LTMs by definition. However, let us propose four desiderata specific to LTM design, that are implied for ``large-scale'' and ``adaptability'' to be possible in tabular data. These are not strictly necessary, but desirable for adapting the model ``to a wide range of downstream tasks'' (see Section \ref{sec:applications}).
% \textcolor{red}{Make slightly more specific to be about generation?}
\begin{enumerate}[label=\textbf{D\arabic*}]
    \itemsep0em
    \item \label{d:mixed}\textbf{Mixed-type Columns.} LTMs should be able to handle the different types of data that are common in tables, e.g. numerical, categorical, datetime, and missing data. 
    \item \label{d:crossdataset}\textbf{Cross-dataset Modelling.} Tabular datasets only cover a specific topic and are often moderately sized. Consequently, to enable large-scale training and broad applicability we need one LTM to be trained on different datasets. This requires a capability to model heterogeneous feature spaces---i.e. feature spaces with different (numbers of) features that could carry very different meanings.
    \item \label{d:context}\textbf{Textual Context.} The meaning of tabular data is often dependent on contextual metadata, e.g. the dataset description, column names, and category names. An LTM should leverage this information. 
    \item \label{d:invariance}\textbf{Invariance/Equivariance w.r.t. column order.} Column order is usually arbitrary for tables. We want an LTM to reflect this, and have output that is invariant or equivariant to input permutations. In other words, letting $f$ denote the LTM, $x$ a row, and $T$ some permutation of the columns, we desire (invariance) $f(T(x))=f(x)$ or (equivariance) $f(T(x))=T(f(x))$.\footnote{Note, equivariance only makes sense if the output is a sequence of the same length as the input. This may not be the case for some LTMs, e.g. LTMs that aim to embed rows as static embeddings.}
\end{enumerate}
% \textcolor{red}{attention-based model is strong contender } 
% As we will see in the next section, current work related to LTMs does not satisfy most of these desiderata.

\parag{Model type and architecture.} Similar to \textit{FM} and \textit{LLM}, we envision the term \textit{LTM} to be very much open to interpretation. In particular, we set no restrictions on the type of models used for LTMs. In LLM literature, conditional generative models with a transformer backbone form a cornerstone for many applications. Yet, it would be false to state encoder-only LLMs that create embeddings are not foundation models. Similarly, we believe some type of conditional generative model (i.e. that allows generation based on a subset of the variables and metadata) could be an important model class for future LTMs: this would for example allow few-shot (probabilistic) prediction, imputation, and conditional generation. For representation learning, encoder-only LTMs may be more suitable. Architecture-wise, attention-based architectures are dominant in related work (Section \ref{sec:related}), for good reason: they can process a varying number of features and naturally satisfy \ref{d:invariance} (i.e. output is equivariant w.r.t. feature order when no positional encoding is used). Nonetheless, benchmarking papers \cite{gorishniy_revisiting_2021,borisov_deep_2022, shwartz-ziv_tabular_2022,grinsztajn_why_2022} could motivate researchers to look beyond transformer architectures.

\subsection{Data} 
Key to FMs, is that they are ``trained on broad data'' \citep{bommasani_opportunities_2022}. In particular, many of the performance gains of recent LLMs and vision models are due to scaling  \cite{brown_language_2020,kaplan_scaling_2020, rombach_high-resolution_2022, hoffmann_empirical_2022}. ``Broad data'' for LTMs implies having a large corpus of tables with context (e.g. descriptions). Fortunately, there has been a growing body of work composing these large datasets.  WebTables \cite{lehmberg_large_2016} consists of 10M HTML datasets from the web. \citet{bhagavatula_tabel_2015} proposed a dataset based on Wikipedia tables, which has the advantage of being curated and covering a wide range of topics. More recently, the size (and hence coverage) of these datasets has increased tremendously. Most recent is \citet{eggert_tablib_2023}, who publish TabLib---a metadataset counting 627 million tables and 827 billion context tokens, covering a diverse range of topics. This is comparable to the size of large-scale image and text datasets, and hence should not prohibit research into LTMs. Nonetheless, there are some possible data challenges, which we will go into in Section \ref{sec:challenges}.

\subsection{Benchmarking tasks for LTMs}
The broad definition of foundation model poses the question: even if we have build a supposed LTM, how do we benchmark it and deem it successful? Multiple tasks, settings, and datasets will likely be required, similar to LLM benchmarks---e.g. the massive text embedding benchmark \cite{muennighoff_mteb_2023} consists of 8 tasks, 56 datasets, and 112 languages. In Appendix \ref{app:benchmarks} we outline some tasks and settings that could be used for benchmarking LTMs. We hope to encourage others to focus on developing metrics and benchmarks further.

\takeaway{We use Large Tabular Model (LTM) to denote a foundation model for tabular data. This term is only loosely defined, but implies a large-scale tabular model that is adaptable to many downstream tasks. }

\section{Current State of LTMs} \label{sec:related}
In the previous section we have discussed our position on what LTMs should be able to do, which included handling some of the inherent difficulties of tabular data (e.g. its lack of structural meaning, mixed-type variables, limited size of datasets).
Let us discuss some of the works related to LTMs, how they already fit some of the LTM criteria, and which criteria they not yet fulfill.

\parag{Representation learning.} 
Both \citet{Yin2020TaBERT} and \citet{deng_turl_2020} adapt BERT \cite{devlin_bert_2019} for table understanding and parsing (e.g. entity linking, column type annotation). Both set-ups are similar, to linearize tables into ``sentences'', add textual metadata, use masked reconstruction loss as training objective, and train on large tabular metadatasets. Their methods have numerous technical limitations that prohibit wide adaptation, e.g. a 128 token context window.
%---600K+ Wikitables \citep{bhagavatula_tabel_2015} and (in the case of \cite{Yin2020TaBERT}) WebTables \cite{lehmberg_large_2016}. 
\cite{zhu2023XTAB} focuses on cross-table pretraining of larger transformer models. They tokenize numerical values more efficiently, as single tokens \cite{gorishniy_revisiting_2021}, and for categoricals train an embedding for each unique category in all datasets. They pretrain on just 100 datasets, and their category look-up prevents large-scale cross-table training (\ref{d:crossdataset}). 
\cite{Ye2023CT-BERT}'s CT-BERT present a similar approach, but instead of linearizing each row as tokens, they embed the tokens of each categorical feature using a pretrained LLM, pool these embeddings, and pass the feature embeddings to their model. This results in shorter sequences, and the pretrained LLM adds context efficiently (\ref{d:context}). They only consider prediction as downstream task, but do show good few-shot performance. Training is limited to 17k datasets and a maximum of 100 features per dataset, but we expect this could scale to larger models and be adapted beyond prediction.

\parag{Supervised learning}
Though supervised learning methods are usually disqualified from being foundation models, we find three works noteworthy. LIFT \citep{dinh_lift_2022} finetunes an LLM on linearized tabular data, and extensive experiments indicate similar performance to traditional baselines. TabFM \cite{zhang_towards_2023} takes a more general approach, representing rows differently, adding task-specific information, and showing good few-shot prediction using in-context examples. TabPFN \citep{hollmann_tabpfn_2023} takes a vastly different angle than the previous works: it is a transformer model that is trained on large quantities of tabular \emph{synthetic} data, to provide few-shot priors for supervised learning. Despite it being trained on toy data, the authors claim that models trained with TabPFN priors outperform XGBoost \cite{chen_xgboost_2016} on real-world datasets. This generalization capability, from toy to real datasets, hints at there being more structure in the tabular domain than some previous works suggest \cite{grinsztajn_why_2022}. %\citet{wang_unipredict_2023} proposes a method that finetunes an LLM on 169 datasets and is able to predict few-shot on new datasets.

\textbf{Generative learning.}
Generative learning approaches are lacking behind significantly. \citet{yoon_radialgan_2018} present a GAN approach that learns to generate samples across domains using a shared latent space, similar to Cycle-GAN \cite{zhu_unpaired_2017} in the image domain. Their intention is to augment smaller domains with synthetic samples that have been ``translated'' from larger domains, so that this improves supervised learning on these small domains. The problem of this approach is that their method uses a separate encoder-decoder pair for each domain to translate to and from the shared latent space. This will not work well for very small domains, does not scale well to many domains, and prohibits few-shot translation. Additionally, their method does not satisfy \ref{d:mixed}, \ref{d:context} and \ref{d:invariance}. 

Pretrained LLMs may provide a solution to generation, as they are generative, contain general knowledge, and can tokenize and embed rows into sequences of meaningful numerical vectors (\ref{d:context}). However, we believe there are significant disadvantages to adapting LLMs for tabular generation. Let us elaborate.

\subsection{Tabular Generation with LLMs}  \label{sec:llm_bad}
Three recent works adapt LLMs for tabular generation. \citet{borisov_language_2023} use a finetuned LLM out-of-the-box for generating tabular data. \cite{solatorio_realtabformer_2023} propose a similar model based on GPT-2, but which can also generate relational datasets and improve efficiency through modelling a reduced fixed-set vocabulary for each column (following \cite{padhi_tabular_2021}). Last, \cite{zhao_tabula_2023} show they achieve better performance by retraining a smaller LLM. Though these works finetune/train the LLM on just a handful of datasets---probably because their LLM backbone is expensive---they could relatively easily be applied to larger corpi of datasets. The advantage of LLM-based generation is its simplicity, it not requiring any manual preprocessing of data, and it allowing typical LLM techniques (e.g. prompting and in-context examples). However, LLMs have three major shortcomings for generating tabular data. 

\parag{LLMs are not good at modelling continuous variables.}
Standard LLMs are inefficient for tabular data, due to tokenization of numerical values \cite{thawani_representing_2021}. A single numerical variable is implicitly modelled as an autoregressive series of binned, categorical variables (e.g. $1.89$ is modelled as $1\rightarrow .\rightarrow 89$). This lacks any prior on the continuity of common data distributions, e.g. predicting $1.89$ or $1.90$ should have about the same probability, in contrast to $1.89$ and $90.89$, yet both differ in just one token. This has been shown to lead to unrealistic outputs in LLMs \cite{spithourakis_numeracy_2018}. 

Some have aimed to improve numeracy through better tokenization of numerical values \cite{spithourakis_numeracy_2018, jiang_learning_2020, golkar_xval_2023}. Though we consider this promising, current solutions come with side effects. For example, some encode number $r$ into fixed vector $r\mathbf{e}$ with $\mathbf{e}$ some constant vector, but this results in numerical variables being modelled as point estimates---removing the ability to generate using the LLM.

\begin{figure}[hbt]
\vspace{-3mm}
    \centering
    \includegraphics[width=1.0\columnwidth]{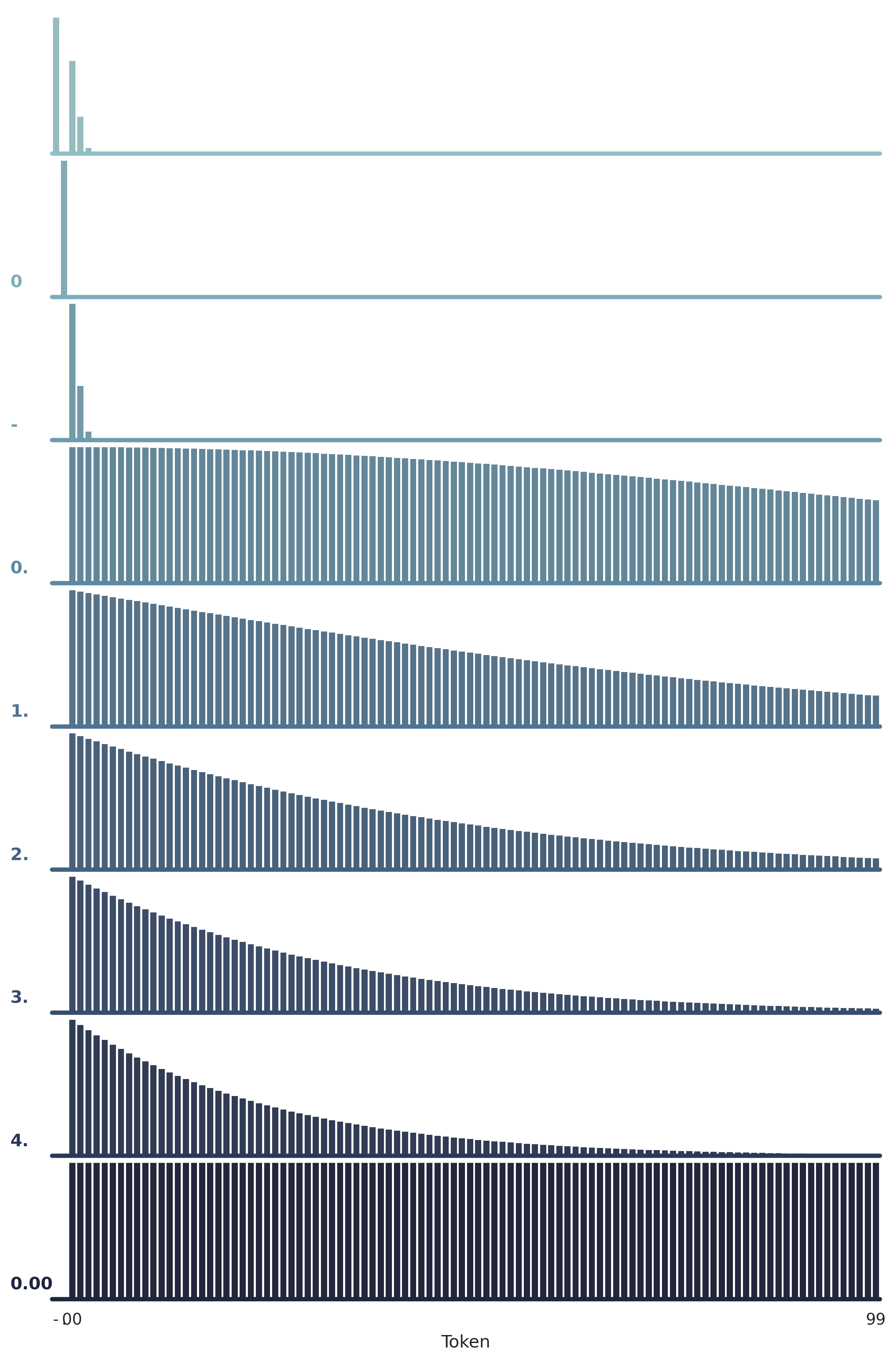}
    \vspace{-3mm}
    \caption{\textbf{Sampling continuous distributions using LLMs autoregressively is inefficient}. Assume we autoregressively sample tokens aiming to generate numbers that follow a standard Gaussian. What token probabilities should the LLM output at each sampling step? Let us consider a total vocabulary of just 102 tokens, $[``-", ``.", ``00",...,``99"]$. For different direct histories of generated text (examples given by each row, already generated digits on left), the output probabilities need to be very different. For example, a single ``draw'' is generated by sampling from probabilities in the first row (e.g. giving``0''), then second row (conditional on ``0'', giving ``.''), fourth row (``0.''$\rightarrow$ e.g. ``00''), and last row (``0.00''$\rightarrow$ e.g. ``00'').}
    \label{fig:binning}
    \vspace{-2mm}
\end{figure}

This is even more problematic for generative tasks in which we aim to sample from the LLM to mimic a continuous distribution. In Figure \ref{fig:binning} we visualize the relatively complex process for an LLM to output just a single independent Gaussian variable. Evidently, modelling just a simple continuous variable requires a seemingly unnecessary amount of capacity, without even considering conditional dependencies on other variables. Other model classes (e.g. diffusion models \cite{song_generative_2019,ho_denoising_2020}) inherently support continuous variables, which we believe makes them a better option.

\parag{LLMs are poorly calibrated.}
At the same time, \citet{renda_can_2023} found that LLMs are poor at sampling from theoretically trivial distributions, e.g. a uniform distribution. This is in line with the poor calibration of LLMs \cite{desai_calibration_2020, jiang_how_2021, jiang_generative_2023}. This does not bode well for modelling more complex, continuous distributions autoregressively.

\parag{Standard LLMs are expensive.}
LLMs are designed to be highly flexible, and typically function autoregressively. This flexibility means we may learn difficult relationships (e.g. model continuous distributions, Figure \ref{fig:binning}). For tables, where a single row may contain hundreds of features that each contain several tokens (especially if numbers are tokenized naively), this would mean inefficient, slow, and expensive training and inference---lessening research and adaptation. For large datasets or inference with in-context examples, the context-window size could also be a hard limitation, e.g. GReaT \cite{borisov_language_2023} cannot handle more than 100, low-dimensional in-context samples.
But is this expense necessary? \citet{Ye2023CT-BERT} use a pretrained LLM for encoding fields, but use a smaller transformer on top---this approach reduces the sequence length to the number of features (cf. tokenized length of row and context) and speeds up training significantly through caching. Furthermore, smaller transformers may suffice: tabular data is highly structured and current state-of-the-art methods, e.g. XGBoost for prediction, perform well despite their small size. Consequently, we believe it would be fruitful to explore other architectures (including transformer-based) beyond the standard LLM.

\takeaway{A recent surge of papers use cross-table training for unsupervised or supervised learning. Many adapt LLMs, but we argue why LLMs are not efficient and performant in the presence of numerical columns. Building a generative LTM is the largest challenge.}

\section{Real-World Impact} \label{sec:applications}
\subsection{Categorizing Adaptation}
The foundation of machine learning (ML) is data. Unfortunately, data can be limiting for some domains (e.g. healthcare, where data is costly and privacy-sensitive), subpopulations (e.g. historically underrepresented minorities), and regions (e.g. developing countries). A lack of good data hinders the development, training, and testing of ML models, and science in general. In particular, we have discussed how ML for tabular data is lagging behind, which may be explained by modern ML methods requiring large training sets that the tabular domain does not always offer (Section \ref{sec:introduction}). Additionally, even if data is available, it may not always be easy to derive meaning or knowledge from it. LTMs can help: LTMs may exhibit the same few-shot abilities as LLMs \cite{brown_language_2020}, allowing us to adapt them to a wide variety of downstream tasks with fewer data. We highlight some applications for this adaptation.

\parag{Direct vs indirect adaptation.} We consider two types of adaptation: direct adaptation to the target task (e.g. fine-tuning the LTM for some prediction task) or indirect adaptation through synthetic data (e.g. create a ``fake'' dataset that augments the real data). To illustrate the advantages of each, let us consider classification on a small imbalanced dataset as downstream task.\footnote{There are strong connections to the debate around discriminative versus generative machine learning models. However, FMs are usually self-supervised (and many FMs like LLMs are generative by nature), hence adapting an LTM to be generative may not necessarily be harder than making it discriminative.} For direct adaptation, one can fine-tune on this small dataset or supply the data as in-context examples, such that the LTM itself provides predictions. This directness is convenient, could be cheap, and allows us to measure the LTM's success directly through its downstream test performance. The indirect approach is different: it uses the LTM to generate a synthetic dataset that augments the real data, which is then used by a downstream model for training. This might be more complex, partly because it is hard to determine how well the LTM did at generating the data until we have run the downstream model (see Section \ref{sec:challenges}). The advantages, however, are that we can (i) explicitly \emph{improve} the real data, e.g. by sampling additional points for the underrepresented group, (ii) can share the synthetic data with others, and (iii) downstream researchers can perform their usual task (e.g. train a model, conduct data analysis), without any need of the LTM. To ensure downstream results are trustworthy, evaluation of the LTM and synthetic data are essential---we discuss this further in Section \ref{sec:challenges}. The choice for direct or indirect adaptation will likely be application-dependent, so let us look into some of these next.\footnote{The ``indirect'' approach is rather unique to tabular data. In the text domain, creating synthetic data\emph{sets} for downstream models often makes little sense---we would still need an LLM to use this data. In the vision generative model literature this is slightly more explored, e.g. \cite{wang_learning_2019} use generative models to turn CGI-generated images into photorealistic training images for better downstream prediction.}

\subsection{LTMs for Responsible AI}
\parag{Inclusiveness and representation}. Historically, it has been hard to model underrepresented groups well due to the inherent data scarcity of these groups. Studying \emph{subgroup robustness} is thus an active area of research, e.g. see \cite{gardner_subgroup_2022}. Large language models have few-shot reasoning abilities \cite{brown_language_2020}, hence we can expect similar behaviour for LTMs. This would allow adapting these models better to small datasets or underrepresented subgroups within large datasets, hence improving performance on these subgroups. Additionally, an LTM could be used to generate synthetic data for these groups---augmenting the real data to improve representation of marginalized groups---in turn enabling better downstream science and ML development for these groups \cite{van_breugel_beyond_2023}. This is similar to other tabular augmentation approaches, e.g. SMOTE \cite{chawla_smote_2002} or the deep generative \cite{van_breugel_can_2023}, but may provide more realistic examples due to the LTM's prior knowledge.

\parag{Robustness through out-of-domain simulation.} ML models can perform unpredictably bad on out-of-distribution and distributionally shifted test data \cite{kolesnikov_wild-tab_2023, liu_need_2023,gardner_benchmarking_2024}. Synthetic data generated by generative models has been used in the past to simulate unseen or scarcely seen scenarios for testing. For example,  \citet{van_breugel_can_2023} show how simulated data can be used for estimating the change in performance of trained ML predictive models (due to an environment change), and \citet{tucker_generating_2020} generate synthetic data to assess healthcare software. These methods are heavily limited by their inability to generate out-of-distribution data, and their generative models' data-intensive requirement for training. The generalization and few-shot potential of LTMs could resolve both problems, thus LTMs could play a central role in simulating data for ML model development, probing, and post-deployment monitoring.

\parag{Privacy, data democratization, and reproducibility.} Training ML models on private data requires a trade-off between privacy leakage and performance \cite{dwork_algorithmic_2014}. LTMs that have been pretrained on open, non-private data can be adapted to private target data, thereby reducing the privacy budget compared to standard private ML models---we need less information from the target dataset to model it well. Similarly, an LTM can generate synthetic data with a better privacy-utility trade-off compared to normal generative models. Both direct and indirect adaptation may prove especially beneficial for underrepresented groups, as these groups have inherently a worse privacy-performance trade-off: there are less points to learn from, hence more information per sample is needed for accurate modelling \cite{van_breugel_membership_2023}. 

Generating synthetic data that accurately mimics sensitive data without revealing it, could also allow the publication of more datasets (e.g. data from healthcare providers). This could promote better reproducibility, democratise data access, and enable more powerful meta-analyses (see below).

\subsection{LTMs for Science} \label{sec:science}
\parag{Meta-analyses.} Meta-analyses are essential for consolidating and analyzing research results across studies. Individual studies may only focus on small subpopulations (e.g. one country), resulting in sometimes conflicting, biased, or insignificant results. At the same time, obtaining more data is often difficult and expensive. Meta-analyses re-use data from multiple, existing studies to create more powerful and cost-efficient analyses, removing local bias and noise. They are influential in informing policy---e.g. meta-analyses are at the top of evidence-based practice \cite{haidich_meta-analysis_2010}. 

In meta-analyses, different datasets need to be consolidated into one, which is complicated by possible heterogeneity (e.g. different features and contexts). LTMs could reduce manual effort and cost by automating the integration of datasets from different studies by harmonizing formats, inconsistent column names and categories, and data types. 

\parag{Bridging datasets.} Going even further, LTMs could help find and combine relevant tables that may not match directly, and may come from very different scientific fields. This can be compared to performing seamless, automated SQL joins using the LTM's full training database.%\footnote{In practice, this could be achieved either by letting the LTM}
This could enable multidisciplinary science that is currently intractable. 

\parag{Data scientist's assistant.} 
LTMs' native understanding of data compared to LLMs, could make them a suitable assistant tool for data scientists. Abilities could include automatic cleaning, exploratory data analyses, finding related datasets, applying SQL queries on complex databases, running automated statistics, and helping visualize and interpret results. This could improve data scientist' productivity, enjoyment of their work (by removing repetitive and often frustrating cleaning), and aid in outwards communication----e.g. in improving public understanding of data meaning and reliability.

\parag{Knowledge base.} By learning from a wide range of datasets, the LTM itself becomes a repository of knowledge. LTMs trained on large, high-quality data (e.g. WikiTables) could facilitate data-driven question-answering systems, and statisticians could distill Bayesian priors from LTMs.

\subsection{LTMs and Non-Tabular Data}
\parag{Time-series and relational data.} We have focused on the simplest form of tabular data: single tables with independent entries. This could lay the foundation for time-series and relational databases, which are important data types in practice (e.g. electronic healthcare records), yet contain dependent samples. In time-series, entries are usually timestamped and multiple samples are linked by an ID (e.g. a patient, where entries correspond to measurements), and in relational databases there can be multiple tables where one ID is linked to (possibly many) entries in different tables.

\parag{Multimodality.} This paper's supposed dichotomy between tabular and image/text is false of course. Many domains are multimodal. LTMs can be aligned or used to inform models in other modalities, and vice versa. For example, desideratum \ref{d:context} states LTMs should use textual context---it would be sensible to acquire this context using a pretrained LLM encoder. Similarly, some of the applications above could still benefit from an LLM intermediary to communicate between the LTM and human. Research into LTMs can also be encapsulated into multimodal FMs, which so far largely ignore tabular data \cite{li_multimodal_2023}.

\takeaway{We envision an important role of LTM in promoting responsible AI (including privacy, representation, reproducibility, data democratization, robustness testing and as a powerful tool for scientists (including data preprocessing and analysis, for bridging datasets, and as a source of knowledge itself). LTMs will also play a role for ML research into other modalities, or multimodal settings.}

% Could also include: (1) fairness/representation, e.g. bespoke data generation, (2) foundation model capabilities, (3) LLMs

% Make use of different datasets
% Data generation with new properties, e.g. 
% fairness or better representation
% Imputation
% Few-shot prediction
% Scientific analyses, i.e. getting useful priors
% Privacy, lower privacy budgets

% Current work is lacking
% generative models

\section{Challenges} \label{sec:challenges}
\parag{Building generative LTMs.} We have seen how generative LTMs are lagging behind. Modelling \emph{distributions} across datasets (\ref{d:crossdataset}), types (\ref{d:mixed}) with context (\ref{d:context}) is conceptually and architecturally complicated, and poses unique and exciting challenges for researchers.

\parag{Scale.} As discussed in Section 3, current models are still limited in scale and applicability. Training is usually restricted to relatively small number of datasets (e.g. 100), and often evaluation is restricted to one (supervised) task. As a result, it is very uncertain whether these models generalize well to new datasets and tasks, rendering them unsatisfactory foundation models. Upscaling to larger datasets may be harder than in other modalities, however, because data is ``dirtier'' (e.g. consist of different file formats, contain missing and noisy fields, or be preprocessed).
% Building an LTM with desiderata \ref{d:mixed}-\ref{d:invariance} may prove an exciting challenge technically. However, there are also challenges in terms of data, bias, and evaluation.

\parag{Data diversity and quality.}
The size of datasets like TabLib \cite{eggert_tablib_2023} is impressive, but it is yet to be determined whether their diversity is sufficient for wide applicability of future LTMs trained on this data. In the image and text domain, large-scale datasets composed of web data are likely to cover a large part of the ``world-distribution'', e.g. include photographs from places all over Earth, and text from different languages. Even then, medical images or small languages may be underrepresented. In the tabular domain, underrepresentation is more likely to be problematic for two reasons. First, because tables have heterogeneous feature spaces and context, most tables are irrelevant to most other tables (cf. images, where each image can be scaled to the same space and where common shapes compose objects, which compose scenes, which compose meaning). Secondly, many forms of tabular data are proprietary, unpublished, or private, and will thus not appear in datasets like TabLib. As a result, the extent to which LTMs will generalize to the medical domain is yet to be seen. On the other hand, recent work \cite{gunasekar_textbooks_2023} have shown that small LLMs trained on small, but highly curated data (e.g. textbooks), can perform remarkably well. This is promising for LTMs, especially LTMs specialized to smaller domains for which data banks exist (e.g. Wikipedia tables \cite{bhagavatula_tabel_2015} or gene expressions \cite{clough_gene_2016}).
% - Non-trivial how to implement the above
% - Reliability: Synthetic Data metrics are notoriously sketchy
% - Misuse
% - Bias 

% 
\parag{Evaluation.} Foundation models make mistakes, e.g. hallucinate false information \cite{ji_survey_2023}, hence careful evaluation is essential. \citet{bommasani_opportunities_2022} divide evaluation into intrinsic and extrinsic. Extrinsic evaluation refers to measuring the performance of the adapted FM on a downstream task, whereas intrinsic evaluation refers to directly measuring the FM's quality. Extrinsic evaluation may not be representative of how ``good'' an FM is, as it may not capture the performance on other possible downstream tasks. On the other hand, intrinsic evaluation is difficult as FMs are defined in terms of their adaptability to downstream tasks, which is hard to measure without performing (or even knowing what are) those tasks. Intrinsic metrics can also be biased towards one FM---e.g. using test loss of FMs with different loss functions is not possible. LTM evaluation encounters the same difficulties, but is made even more difficult by \emph{visual inspection being unhelpful}: whereas the realism of image and text output can be relatively easily evaluated by humans (and forms a central part in most text and image papers), this is hard for tables. Inspiration may be drawn from the synthetic data literature, where similar evaluation challenges make metrics an active area of research \cite{van_breugel_beyond_2023}. Until intrinsic metrics are improved, we expect most LTM performance evaluation to be mostly extrinsic, using multiple downstream tasks. In Appendix \ref{app:benchmarks} we include different tasks and settings that could be considered for extrinsic evaluation.

Evaluation goes further than measuring performance. LTMs need to avoid leaking private or copyrighted material, which requires measuring a model's memorization. Ideas from generative model generalization metrics \cite{theis_note_2016, arora_generalization_2017, alaa_how_2022} and privacy-focused ML \cite{dwork_algorithmic_2014} could inspire LTM metrics. Evaluating bias is another challenge, which we discuss separately.

% \parag{Making tabular data research less dry.} The non-human-like output assessing tabular data visually also leads to one metascientific challenge: how do we as researchers 

\parag{Bias.} Similar to LLMs, LTMs may copy or even exacerbate bias in their training data. Not much is known about the bias in large tabular datasets, e.g. \cite{eggert_tablib_2023}. One may hope that tables are usually created to represent information and facts, and are less opinionated than text. This would be naive: datasets can contain discriminative features (e.g. the much-used Boston Housing dataset \cite{harrison_hedonic_1978}), under- or misrepresent some groups, and display other forms of bias, e.g. publication bias \cite{thornton_publication_2000}. Research into LTMs should go hand-in-hand with research into this bias, and put safeguards in place when publishing LTMs.

\takeaway{The most important challenges for LTMs relate to (i) the challenges of building a model that can handle the unique challenges of tabular data (\ref{d:mixed}-\ref{d:invariance}); (ii) uncertainty around current datasets' quality, bias and diversity; (iii) reliable evaluation of LTMs; and (iv) how bias can be tested and safeguarded against in trained LTMs.}

\section{Comparing LTM and LLM Impact} \label{sec:comparison}
Let us end this position piece with a comparison of LTMs and LLMs along five key dimensions of impact. 

\parag{1. Public and commercial use.} The surge in LLM use is a testament to their adaptability, and the fundamental role they may play in productivity and facilitating human-ML interactions (e.g. through tools called through LLMs). LTMs may play an important role in data-driven industries, e.g. finance, education, government, but this will likely be smaller than LLMs. 

\parag{2. Scientific use.} Tabular data is likely the largest data modality in science \cite{borisov_deep_2022}. We have discussed how LTMs may revolutionize how tabular data is processed, analyzed, and re-used across domains. Though LLM applications in the scientific domain are plentiful, current LLMs encounter inefficiency and numerical issues with tabular data (Section \ref{sec:llm_bad}).

\parag{3. Potential for public good.} LLMs may play a central role in education,  personalized healthcare \cite{gates_age_2023}, and law \cite{bommasani_opportunities_2022}. LTMs impact is much more specific to ML and data science, e.g. more inclusive data, the development of more robust models, improving privacy and reproducibility, and scientific knowledge derived from using LTMs (e.g. that use multidisciplinary data). Consequently, both exhibit great potential for public good, yet cover vastly different areas.

\parag{4. Risk of misuse.} Generative FMs come with a risk of misuse. Highly-realistic content generated by LLMs or vision models can be used to impersonate individuals (i.e. deepfakes) \cite{rana_deepfake_2022}, spread misinformation at an unprecedented scale \cite{marcus_why_2023}, and raise plagiarism concerns for creators and academia \cite{kasneci_chatgpt_2023}. This risk is smaller for LTMs compared to text and vision, because contributing tabular data to the community often requires identity verification (i.e. data can be traced back to an individual and institution) and is prone to scrutiny (i.e. data scientists are strained to verify the data source). Additionally, malicious data fabrication to fit a false narrative can already be done easily manually without the need of an LTM (cf. image generators, where the alternative is arduous and skillful photoediting). FM risks can also be unintentional however, e.g. it is difficult to guarantee models are unbiased, reliable, and do not reproduce copyrighted or private information. For evaluation of these unintentional consequences, LTMs and LLMs face similar challenges.

\parag{5. Impact per researcher.} The amount of research in the LLM space vastly outnumbers work related to LTMs, yet there are many interesting, high-potential open questions and applications for LTMs. Furthermore, the scale of current LLM research requires vast computational resources, whereas LTMs have been underexplored at even moderate scale. Consequently, the potential for impact is arguably more plausible and more widely attainable in LTM research.

\takeaway{The above is \textit{not} to give a verdict on which modality is best---inherently we are comparing apples to pears. However, we do hope to make the point that each model class is very much deserving of researchers' attention---attention that LTMs are currently not getting.}

\section{Conclusion} 
We believe large tabular models are significantly understudied and within reach. LTMs provide exciting and unique research challenges, impactful applications, and promises a world where data usability extends across single datasets. We hope some readers will reconsider the modality they prioritize in their research, and help build, evaluate, and responsibly apply the first generation of true LTMs.

\section*{Impact Statement} In this work we have argued for the importance of more research into tabular foundation models. In Section \ref{sec:applications} we have elaborated on how these methods can aid ML inclusiveness, representation, privacy, data democratization, and robustness. On the other hand, we have also discussed how LTMs, like LLMs, can be biased, misused, and require proper evaluation (Section \ref{sec:challenges}). We acknowledge that research on and adaptation of LTMs need to go hand-in-hand with bias and misuse prevention.

\clearpage
\section*{Acknowledgements}
This research has been supported by the Office of Naval Research UK. Additionally, we would like to thank Julianna Piskorz for some very useful feedback on the initial draft of this paper, and we thank the four ICML reviewers for their constructive comments---all of this has greatly improved the paper. Lastly, BvB is grateful for the brainstorming input and continuing support of Sophie Mary.

\bibliography{references_zotero}
\bibliographystyle{icml2024}

%%%%%%%%%%%%%%%%%%%%%%%%%%%%%%%%%%%%%%%%%%%%%%%%%%%%%%%%%%%%%%%%%%%%%%%%%%%%%%%
%%%%%%%%%%%%%%%%%%%%%%%%%%%%%%%%%%%%%%%%%%%%%%%%%%%%%%%%%%%%%%%%%%%%%%%%%%%%%%%
% APPENDIX
%%%%%%%%%%%%%%%%%%%%%%%%%%%%%%%%%%%%%%%%%%%%%%%%%%%%%%%%%%%%%%%%%%%%%%%%%%%%%%%
%%%%%%%%%%%%%%%%%%%%%%%%%%%%%%%%%%%%%%%%%%%%%%%%%%%%%%%%%%%%%%%%%%%%%%%%%%%%%%%

\newpage
\onecolumn
\appendix
\section{Details Figure 1} \label{app:figure_details}
The rough estimates in Figure 1 are created by checking the titles and abstracts of all accepted papers on the presence of certain keywords, and plotting the number of counts per modality which describe ``foundation model" in some way. The keywords for ``foundation model'' and each modality are given in Table \ref{tab:keywords}. Papers with multiple modalities are counted towards each modality. The total number of accepted papers per conference are 1828 (ICML 2023), 3218 (NeurIPS 2023), and 2250 (ICLR 2024).

\begin{table}[hbt]
    \centering
    \caption{Key words for each search term. Abstracts are made lower-case. We intentionally do not include ``text'' and ``time'' as keywords for respectively Language and Time Series applications, as we found this lead to many false positives. $^*$ For ``table'' and ``llm'', we use a word boundary ($\backslash b$) at the start of the keyword, to avoid common false positives (e.g. ``portable'', ``mutable'', ``bellman'').}
    \begin{tabular}{c|c}
    \toprule
         Research Topic & Keywords  \\ \midrule
         Foundation Model & foundation model, llm$^*$, cross-dataset \\ \hline
         Graph & graph \\
         Image & image, vision \\
         Language & language, llm \\
         Table & tabular, table$^*$, cross-table \\
         Time Series & time-series, time series, temporal \\  \bottomrule
    \end{tabular}
    \label{tab:keywords}
\end{table}

Papers that satisfy both ``foundation model" and ``table" and thus form ``Table" group in Figure \ref{fig:conference_stats} are the following per conference: ICML 2023 \cite{zhu2023XTAB,ni_lever_2023}, NeurIPS 2023 \cite{li_sheetcopilot_2023,zhu_towards_2023,ajay_compositional_2023,wang_describe_2023,hollmann_large_2023,manikandan_language_2023}, ICLR 2024 \cite{li_chain--knowledge_2024,wang_chain--table_2024,patnaik_cabinet_2024,kong_opentab_2024,yuan_craft_2024,bao_fast-detectgpt_2024,yang_unitabe_2023}. We note these include LLM works that are evaluated on tabular tasks, e.g. Table QA \cite{ni_lever_2023}.

\section{Benchmarking LTMs} \label{app:benchmarks}
For LTMs, benchmarks will be very important, yet the breadth of their applications will make it hard to fairly compare models. More research into LTM benchmarks is essential. For now, we would like to indicate a number of relevant benchmarks. We split this up in ML tasks and experimental set-ups, where the latter aims to measure adaptability of the LTM for different data settings.

\textbf{Tasks} could include: 
\begin{enumerate}
    \item \textbf{Supervised learning}. Predictive performance can be measured either through using the LTM for retrieving row representations that can be used by a small downstream predictor, or by using the LTM directly (i.e. by generating the target conditional on features). 
    \item \textbf{Synthetic data generation}. For generative LTMs, conditional and unconditional generation quality can be measured similar to more traditional generative models. Any synthetic data metric can be used, e.g. train-on-synthetic-test-on-real performance (for downstream utility) \cite{jordon_synthetic_2022}, fidelity and diversity metrics \cite{sajjadi_assessing_2018,kynkaanniemi_improved_2019,naeem_reliable_2020}, and $\epsilon$-identifiability (for privacy) \cite{yoon_anonymization_2020}.
    \item \textbf{Imputation}. Generative LTMs can be used for imputation, and benchmarked on this. This can be further split up in single imputation tasks (where the aim is to model $\mathbb{E}(X_{unobserved}|X_{observed})$) or multiple imputation (where the aim is to sample from $p(X_{unobserved}|X_{observed})$). The latter is only relevant for generative LTMs (as representation learning-based LTMs only provide point estimates). Another axes could be the missingness mechanism (MCAR, MAR, MNAR).
    \item \textbf{LTMs for science.} This includes a range of tasks that require more research and adaptation beyond the current line of LTM work, but which could help describe success in some of the applications described in Section \ref{sec:science}. For example, subtasks could include dimensionality reduction, data cleaning (e.g. outlier detection), clustering, and cross-dataset tasks (e.g. ``which of the following datasets is most relevant to [some other dataset or question]").
\end{enumerate}

\textbf{Experimental set-ups} may include:
\begin{enumerate} 
\item \textbf{Few-shot.} Performance on new datasets for which relatively few samples are available. This could be split up further into approaches that allow for finetuning of the LTM, and methods that do not (e.g. in-context learning for LLMs). For this setting, both performance and robustness of the LTM (compared to baseline methods trained on the same few samples) could be a deciding factor in measuring the LTMs success. This closely relates to out-of-distribution and distributional shift benchmarking, for which good benchmarks exist \cite{kolesnikov_wild-tab_2023,gardner_benchmarking_2024}.
\item \textbf{Zero-shot.} This is the same as few-shot, but without any samples from the target dataset. This requires true generalization of the LTM, and is likely not achieved by the current LTMs.
\item \textbf{In-distribution.} The performance of an LTM on hold-out test sets from the datasets on which the LTM was trained. This is less interesting for foundation models, as we generally want adaptability---i.e. genereralizability \emph{beyond} the training data. Nonetheless, in-distribution evaluation would be useful to quantify generalization gaps in the previous two settings (e.g. compare few-shot performance w.r.t. in-distribution experiments).
\end{enumerate}

One difficulty in these experimental setups is that pretrained and published foundation models do not always come with descriptions of the precise data (or splits thereof) on which they were trained. This is a ubiquitous problem in foundation model evaluation, and especially problematic when one assumes they are measuring few-shot or zero-shot performance, but actually data leakage has occurred.

\end{document}